\documentclass[runningheads]{llncs}
\usepackage[T1]{fontenc}
\usepackage{graphicx}

\usepackage{subcaption}

\newcommand{\includesvg}[2][]{%
  \includegraphics[#1]{#2_svg-tex.pdf}%
}

\usepackage{caption}
\usepackage{booktabs}
\usepackage{multirow}

\usepackage{threeparttable}

\usepackage{amsmath}
\usepackage{amssymb}

\usepackage{enumitem}

\usepackage{xcolor}
\definecolor{darkgreen}{rgb}{0.0, 0.7, 0.0}

\newcommand{\cmrd}[1]{#1}

\usepackage{url}

\usepackage{fancyvrb}
\usepackage{fvextra}
\usepackage{tabularx}
\usepackage{listings}

\usepackage[most]{tcolorbox}
\tcbset{
  promptbox/.style={
    enhanced,
    breakable,
    colback=white,
    colframe=black!40,
    boxrule=0.6pt,
    arc=2pt,
    left=6pt,right=6pt,top=6pt,bottom=6pt,
    fonttitle=\bfseries,
  }
}

\begin{document}
\title{CP-SynC: Multi-Agent Zero-Shot Constraint Modeling in MiniZinc with Synthesized Checkers}
\titlerunning{Multi-Agent MiniZinc Modeling with Synthesized Checkers}
% If the paper title is too long for the running head, you can set
% an abbreviated paper title here
\author{Yuliang Song \and Eldan Cohen}
\authorrunning{Y. Song and E. Cohen}
\institute{Department of Mechanical and Industrial Engineering, \\ 
           University of Toronto, Toronto, Canada \\
\email{yl.song@mail.utoronto.ca, eldan.cohen@utoronto.ca}}
\maketitle              % typeset the header of the contribution

\begingroup
\renewcommand{\thefootnote}{}
\footnotetext{\raggedright\footnotesize  \cmrd{The code and benchmark used in this paper are available at} \url{https://github.com/Yuliang795/LLMs-CP-CPSynC}.}
\endgroup

\begin{abstract}
Constraint Programming (CP) is a powerful paradigm for solving combinatorial problems, yet translating natural language problem descriptions into executable models remains a significant bottleneck. While Large Language Models (LLMs) show promise in automating this translation, they often struggle with subtle semantic errors in the absence of oracle validation at test time. To address this, we introduce \textsc{CP-SynC} (Constraint Programming modeling with Synthesized Checkers), a multi-agent workflow for zero-shot constraint modeling in MiniZinc.
\textsc{CP-SynC} coordinates modeling agents that generate and refine candidate models and validation agents that synthesize semantic checkers to provide feedback on semantic correctness.
To mitigate noise inherent in individual LLM outputs, \textsc{CP-SynC} explores multiple modeling trajectories in parallel and employs selection agents to select the final model via multi-agent evidence aggregation.
Extensive experiments on a benchmark of 100 CP problems show that \textsc{CP-SynC} \cmrd{substantially} outperforms existing baselines in MiniZinc modeling.

\end{abstract}

\section{Introduction}

Combinatorial decision problems are pervasive in modern industrial systems, underpinning scheduling, transportation, and production planning \cite{rossi2006handbook,petropoulos2024operational}.
Constraint programming (CP) provides a powerful paradigm for addressing such problems by translating them into formal constraint models, where a solver then automatically searches for solutions \cite{rossi2006handbook}.
However, translating natural-language requirements into precise, machine readable constraint models remains a demanding manual process that requires substantial constraint programming expertise beyond domain knowledge \cite{akgun2011extensiblecp,michailidis2024constraint,michailidis2025cpbench,szeider2025cpagent}.
Despite remarkable advances in solver technology, this modeling bottleneck continues to restrict the wider adoption of CP, motivating research on methods that enable domain experts to address a broader range of problems without deep CP expertise \cite{o2010automated}.

The rapidly advancing coding and reasoning capabilities of Large Language Models (LLMs) \cite{cao2025pragmatic} have sparked significant interest in leveraging them to simplify the modeling process.
In particular, a growing body of research explores developing LLM-based systems for automating the translation of natural language problem requirements into executable CP models 
\cite{michailidis2024constraint,song2025llmcp,singirikonda2025text2zinc,szeider2025mcp}, thereby lowering the entry barrier for practitioners.
While recent works have demonstrated the feasibility of this translation, they also reveal critical challenges, both syntactic and semantic \cite{song2025llmcp,michailidis2025cpbench}. 
Although LLMs often struggle to generate syntactically valid CP models in one shot, syntax errors are reliably reported by the compiler and can be effectively repaired by LLMs \cite{song2025llmcp}.
In contrast, semantic errors remain the primary obstacle as they are difficult to identify without oracle validation, and can be more detrimental in practice if left unrecognized.

In this work, we introduce \textsc{CP-SynC} (Constraint Programming modeling with Synthesized Checkers), a multi-agent workflow designed to improve zero-shot modeling performance. \textsc{CP-SynC} coordinates three complementary roles: (i) modeling agents that generate and refine candidate models; (ii) validation agents that synthesize semantic checkers as auxiliary functions to assess the semantic correctness of a solution; and (iii) selection agents that aggregate evidence across candidate models and their checker outcomes to select the final model. 
During generation, LLM outputs can be noisy or inconsistent due to their probabilistic nature, and relying on a single generation may lead to degraded modeling performance.
To enhance robustness, we generate diverse responses within each role and explore multiple modeling trajectories in parallel, increasing the likelihood of developing a semantically correct model.
In the absence of oracle validation at test time, we use synthesized semantic checkers as approximate supervision for refining candidate models. However, LLM\cmrd{-}synthesized checkers can be unreliable, and naively revising models to satisfy noisy feedback can steer generation toward semantic misalignment \cite{chen2025selftest}. 
To mitigate this, \textsc{CP-SynC} manages agents to reason about checker validity, empowering the workflow to identify and reject erroneous feedback rather than blindly refining against it.
To comprehensively assess our workflow, we introduce an automated benchmark of 100 CP problems across academic and industrial domains.

Our contributions are summarized as follows:
\begin{enumerate}
    \item We propose \textsc{CP-SynC}, a novel multi-agent framework for zero-shot MiniZinc modeling. 
    \textsc{CP-SynC} coordinates complementary agent roles for modeling, validation, and selection, enabling interactive test-driven development with synthesized semantic checkers and robust final model selection through multi-agent evidence aggregation.  

    \item We conduct extensive experiments across five prominent LLMs and representative baselines on an automated benchmark of 100 CP problems from diverse domains.
    Empirical results show that \textsc{CP-SynC} achieves the strongest MiniZinc modeling performance against baselines across all evaluated LLMs.  

    \item Our ablation analysis quantifies the impact of each agent type in \textsc{CP-SynC}, analyzes sampling and selection strategies for multi-trajectory modeling, and characterizes performance gains as the inference budget increases.
\end{enumerate}

\section{Preliminary}
\subsubsection{Constraint Programming}
CP is a powerful paradigm for solving combinatorial problems. The resolution of a constraint satisfaction problem (CSP) typically involves two stages: (i) modeling, where problem requirements are expressed as constraints over decision variables; and (ii) solving, where a solver searches the space of assignments to find feasible solutions subject to the constraints~\cite{rossi2006handbook}. Formally, a CSP model can be defined as a tuple \(\mathcal{M}=(X, D, C)\), where 
\(X=\{x_i\}_{i=1}^n\) is the set of decision variables, 
\(D=\{D_i\}_{i=1}^n\) is the set of corresponding domains, and 
\(C=\{c_j\}_{j=1}^m\) is the set of constraints that must hold among variables.
The goal of constraint solving is to find an assignment \(s = (x_1 = v_1, \dots, x_n = v_n)\)
such that all constraints in $C$ are satisfied.

Constraint Optimization Problems (COPs) extend CSPs with an objective function \mbox{\(f\colon\prod_{i=1}^n D_i \to \mathbb{R}\)} to be minimized or maximized, yielding the tuple \(\mathcal{M} = (X, D, C, f)\).
A feasible solution to a COP satisfies all constraints and an optimal solution is defined as 
\(s^* = \arg\min_{s \models C} f(s)\) for minimization or \(s^* = \arg\max_{s \models C} f(s)\) for maximization.

\subsubsection{MiniZinc}
Automatically solving a CP model requires expressing it as an executable formulation that a solver can process.
MiniZinc serves this role as an open-source \textit{constraint modeling} language for expressing CSP and COP formulations at a high level~\cite{nethercote2007minizinc}. Unlike imperative programming, where the user explicitly dictates detailed steps to find a solution, MiniZinc supports a declarative paradigm in which the user encodes problem requirements into decision variables, constraints, and optionally an objective function. The resulting model is then compiled into a solver-independent format and can be automatically solved by a range of back-end solvers including constraint programming (CP), mixed-integer programming (MIP), and boolean satisfiability (SAT) solvers. 
MiniZinc further offers a broad library of global constraints, which encapsulate recurring combinatorial patterns as high-level primitives. Using global constraints allows modelers to express common structures succinctly and often enables solvers to exploit specialized propagation for better efficiency \cite{nethercote2007minizinc}.
See Appendix~\ref{asec:eg_nqueens} for example MiniZinc models of the N-Queens problem.

\subsubsection{Large Language Models}
LLMs are autoregressive models that parameterize a probability distribution over token sequences, and are typically trained on large text corpora. Given an input sequence \(x_{1:n} = (x_1,\dots, x_n)\), an LLM with parameters \(\theta\) models
\mbox{\(
p_\theta(x_{1:n}) = \prod_{i=1}^{n} p_\theta(x_i \mid x_{<i})
\)}. At inference time, for each generation step \(i\), the model processes the context \(x_{<i}\) and outputs a vector of logits \(z_i \in \mathbb{R}^{|\mathcal{V}|}\) over the vocabulary \(\mathcal{V}\). A temperature-scaled softmax yields the next-token distribution
\[
p_{\theta,\tau}(x_i=v \mid x_{<i}) \;=\; \frac{\exp(z_{i,v}/\tau)}{\sum_{u \in \mathcal{V}} \exp(z_{i,u}/\tau)},
\]
where $\tau$ controls the sampling diversity: lower $\tau$  produces a sharper distribution and more deterministic outputs, while higher $\tau$ flattens the distribution and increases generation diversity \cite{holtzman2019sampling}.

\subsubsection{Prompt Engineering} Prompts are natural-language instructions that condition an LLM's generation for a given task. In a \textit{zero-shot} setting, the LLM receives only the task instruction and relies on its intrinsic knowledge to generate a solution, whereas \textit{few-shot} methods augment the instructions with a set of solved exemplars at inference time to guide generation \cite{brown2020language}. \cmrd{Despite recent success across various applications and domains \cite{chen2025efficient,zhao2025variational}}, LLMs are inherently probabilistic, and generating a correct solution to a complex task in a single attempt remains challenging \cite{madaan2023selfref}. To mitigate this, recent works employ \textit{self-improvement}, where an initial solution is iteratively refined using feedback such as human inspection \cite{chidambaram2024socratic} and ground-truth unit tests \cite{chen2023teaching}.
When such oracles are unavailable, LLMs can instead be prompted to perform \textit{self-checking} as approximate supervision, e.g., by synthesizing checking code to validate a candidate solution \cite{szeider2025cpagent}, or by soliciting self-critiques where the LLM inspects the solution and provides feedback for refinement \cite{michailidis2025cpbench,szeider2025mcp}.

\subsubsection{Problem Definition}
\begin{figure*}[b!] % or [htbp]
    \centering
    \includesvg[width=\linewidth,keepaspectratio]{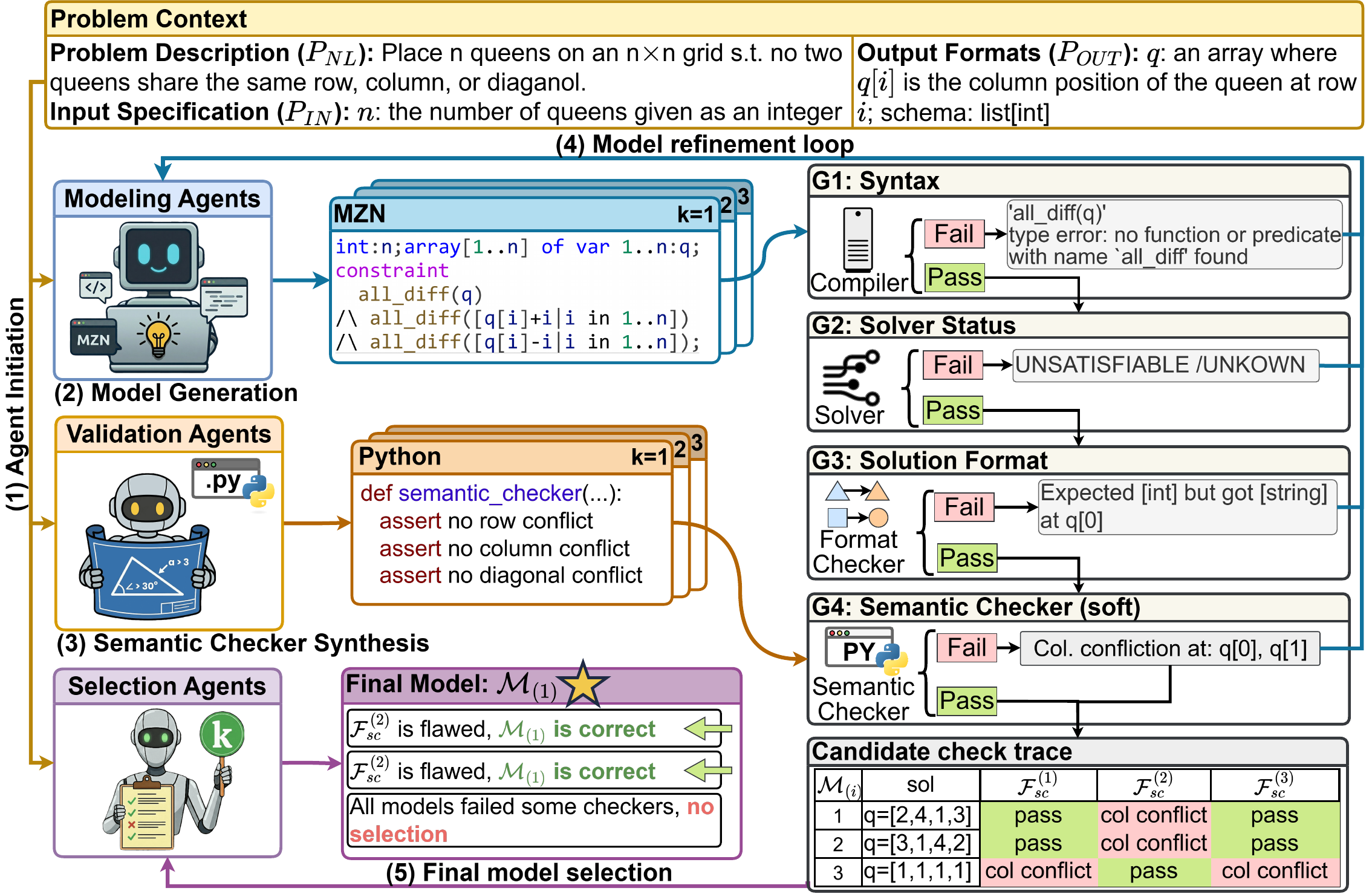}
    \caption{\textbf{CP-SynC workflow overview.} Given a problem context, the workflow proceeds in \cmrd{five steps}: \cmrd{(1) Agent initiation with role-specific system prompts and the problem context; (2) Modeling agents generate $K$ MiniZinc models; (3) Validation agents synthesize $K$ semantic checkers}; (4) Candidates enter the staged checking pipeline (G1--G4), where modeling agents initiate refinements in response to error messages; (5) Selection agents review the candidate models along with their checker outcomes and the semantic checkers, then select the final model by majority vote. If no valid selection is made, the system restarts from Step~1 with a refined problem description.}
    \label{fig:CP-SynC}
\end{figure*}

We formalize automated CP modeling as the translation of a natural-language problem context into an executable CP model. 
To enable automated evaluation of the generated models, we follow prior work \cite{song2025llmcp,singirikonda2025text2zinc} and require solver outputs to be reported in task-specific formats.
Let $\mathcal{P}$ denote the set of problem instances. Each \cmrd{problem $p \in \mathcal{P}$ is defined by a problem context
$p=(P_{NL},P_{IN},P_{OUT})$}, where 
$P_{NL}$ is the natural language description of the task, 
$P_{IN}$ is the input specification, describing the semantics and schema of the input data, and $P_{OUT}$ is the output specification, detailing the semantics and schema of the required output format (See Fig.~\ref{fig:CP-SynC} for an example). When applicable, each problem instance is accompanied by input data \(\mathcal{I}_p\) that instantiates the parameters specified by \(P_{IN}\) (e.g.,  \(N{=}4\) for the N-Queens problem). 
Given a problem context \(p\), the translation is denoted as a function \(\mathcal{F}(p) = (\mathcal{M}_p, \mathcal{S}_p)\),
where \(\mathcal{M}_p\) is the synthesized CP model and \(\mathcal{S}_p\) is the output solution extracted from solving \(\mathcal{M}_p\) subject to \(P_{OUT}\).

\section{CP-SynC}

We present \textsc{CP-SynC} (Constraint Programming modeling with Synthesized Checkers), a multi-agent workflow for end-to-end constraint modeling. Given a problem context $p=(P_{NL},P_{IN},P_{OUT})$, \textsc{CP-SynC} produces a model in MiniZinc and outputs solutions in the prescribed output format.

\subsection{System Design}
\label{sec:system_design}
Conceptually, the workflow is decomposed into three functional components for modeling and refinement, validation, and selection, each implemented by a corresponding agent role as described in \S~\ref{sec:imp_detail} \emph{Implementation Details}. All prompts are provided in Appendix~\ref{asec:prompts}.
Since LLMs often cannot guarantee a correct solution in a single attempt on complex tasks, we employ iterative self-refinement to enhance the quality of individual modeling trajectories, orchestrated by a staged checking pipeline. 
Specifically, the self-refinement process is guided by comprehensive diagnostic feedback spanning the full modeling cycle, including three execution level checks (G1: Syntax, G2: Solver Status, G3: Output Format) and a semantic check (G4), as detailed in \S~\ref{sec:checking_pipeline}. The execution checks act as hard gates that a candidate must pass to proceed, while the semantic check serves as a soft gate whose feedback is used as guidance for modeling and selection. 

\subsubsection{Diverse Sampling Strategies}
\label{sec:sampling_strategies}
During generation, LLM outputs can be noisy or inconsistent due to their probabilistic nature, and relying on a single generation may lead to degraded modeling performance. To enhance robustness, we instantiate multiple agents for each role to produce diverse responses. Specifically, we consider two strategies:
First, we apply \emph{diverse prompt sampling}, which induces diversity by injecting variation into the problem descriptions while keeping the role-specific system prompts fixed. Agents in the same role share the same system prompt but generate conditioned on different problem descriptions, using temperature \(\tau{=}0\). Here, we consider two types of problem description variation: (i) a refined problem description, where we instruct the LLM to review the problem description and fix any detected logical errors with its intrinsic knowledge; and (ii) a planning-augmented problem description, where we instruct the LLM to generate a stepwise modeling strategy and integrate it into the problem description.
Second, we utilize \emph{temperature sampling}, where all agents use the original problem description and diversity is induced by sampling from the LLM with a non-zero temperature at inference.
Empirically, we found that larger LLMs benefit more from diverse prompt sampling towards end modeling performance, while smaller LLMs perform better with temperature sampling.

\subsubsection{Workflow Overview}
\label{sec:workflow_overview}
We present an overview of the workflow below and in Fig.~\ref{fig:CP-SynC}, with detailed descriptions of each component provided in \S\ref{sec:imp_detail}.
Given an input problem context \(p\), \textsc{CP-SynC} executes the loop in Steps~1--5.
If Step~4 or Step~5 emits an abort signal, the system replaces the current problem description \(P_{NL}\) with a refined variant generated using the diverse prompt sampling strategy, and restarts from Step~1 to explore a divergent trajectory.
We set a restart budget of \(R\).
If all \(R\) restarts are exhausted without any model being selected, the system returns a random candidate model that has passed checks G1--G3.

\begin{enumerate}%[leftmargin=*, itemsep=0pt, parsep=0pt, topsep=0pt, partopsep=0pt]
  \item \textbf{Agent Initiation.} We instantiate $K$ agents for each role, assigning each agent the role-specific system prompt along with the problem context $p$, subject to the selected diverse sampling strategy. Specifically, for diverse prompt sampling, we prepare distinct problem descriptions (original, refined, and planning-augmented) and assign one to each agent within the role. For temperature sampling, all agents use the original problem description.
  
  \item \textbf{Candidate model sampling.} A set of $K$ modeling agents each generates one MiniZinc model conditioned on $p$, yielding a portfolio $\{\mathcal{M}^{(k)}\}_{k=1}^{K}$.
  
  \item \textbf{Semantic checker synthesis.} A set of $K$ validation agents each synthesizes one semantic checker conditioned on $p$, yielding a collection $\{\mathcal{F}^{(k)}_{sc}\}_{k=1}^{K}$ .
  
  \item \textbf{Staged refinement loop.} Each candidate model traverses a cascade of three hard execution checks (G1 syntax, G2 solver status, G3 output formats) followed by a soft semantic check (G4) evaluated against all synthesized semantic checkers $\{\mathcal{F}^{(k)}_{sc}\}_{k=1}^{K}$. For G1--G3, the first failure triggers the modeling agent to perform a targeted repair using that check's diagnostic signal, repeating up to a per-candidate refinement budget $r$. After each repair, the updated candidate re-enters at G1. At G4, if a majority of semantic checkers reject the candidate, the modeling agent may either accept the feedback and attempt a repair, or reject the feedback and keep the current model. If no candidate passes all execution checks after refinement, the current loop terminates and emits an abort signal.

  \item \textbf{Final model selection.} Candidates that pass execution checks (G1--G3) are forwarded to a set of \(K\) selection agents together with their G4 outcomes. 
  Each selection agent jointly reviews the candidate models, the synthesized semantic checkers, and the full set of checker outcomes, aggregating these signals to either select the model that is most aligned with the problem context or reject all candidates. The final decision is made by majority vote across the selection agents. If no candidate is selected, the current iteration terminates with an abort signal.
\end{enumerate}

\subsection{Implementation Details}
\label{sec:imp_detail}
\subsubsection{Modeling Agents}
Modeling agents form a crucial component of our workflow, handling model generation, refinement, and output formatting.

\noindent\textit{Model generation:} 
Each agent is provided with the problem context and instructed to produce a MiniZinc model that aligns with the problem description. To avoid relying on a single potentially noisy generation, we explore diverse modeling trajectories by sampling up to \(K\) candidate models using the diverse sampling strategies, yielding a portfolio
\(\{\mathcal{M}^{(k)}\}_{k=1}^{K}\).

\paragraph{Output formatting:} 
Following prior work \cite{song2025llmcp}, we separate constraint modeling from solution formatting to reduce task complexity. 
Once a candidate model passes the solver status check G2 and yields a variable assignment, the agent generates an output formatter $F_{p}$, a task-specific Python function that transforms the variable assignment to the required output format $P_{OUT}$, producing the output solution $\mathcal{S}_p$ used for evaluation.

\paragraph{Model refinement:} 
During the staged refinement loop (Step~4 of the workflow), when a failure is detected at an execution check (G1--G3), the modeling agent is instructed to analyze the error message and propose targeted repairs using that check's diagnostic signal. For the semantic checks (G4) synthesized by the validation agents described below, the modeling agent is instructed to review the feedback together with the checker code, consider whether the failure may be due to flaws in the checker itself, and then decide whether to attempt a repair or reject the feedback and keep the current model.

\subsubsection{Validation Agents}
Validation agents synthesize semantic checkers, implemented as Python functions, that validate candidate solutions against the problem requirements. 
To avoid being biased by incorrect models, validation agents synthesize semantic checkers in a specification-based setting based solely on the problem context $p$, without any information about the candidate model.

Given a problem context $p$, a validation agent first derives constraints from the problem description, then encodes them as assertions on the satisfaction of each constraint. The resulting Python functions validate candidate outputs using the input data $\mathcal{I}_p$ and the output solution $\mathcal{S}_p$. When a derived constraint is violated, the checker returns an informative, constraint level error message. For example, in the N-Queens problem, a candidate solution that violates the column constraint may receive an error message such as ``detected two queens on the same column''.
Because LLM-synthesized checkers can be noisy or incomplete, we synthesize multiple semantic checkers to improve robustness and coverage. Specifically, we instantiate \(K\) validation agents, each synthesizing one semantic checker using diverse sampling strategies, yielding a collection denoted as:
\(
\{\mathcal{F}^{(k)}_{sc}\}_{k=1}^{K},\:\;
\mathcal{F}^{(k)}_{sc}(\mathcal{I}_p, \mathcal{S}_p)
   \rightarrow (\texttt{pass/fail},\, \texttt{feedback}) \, .
\)

\subsubsection{Selection Agents}
Selection agents constitute the decision layer of \textsc{CP-SynC} and select the final model in two stages. To mitigate noise from an individual agent, we instantiate \(K\) selection agents using the diverse sampling strategies and aggregate their selections.
\textbf{Stage 1: Multi-Agent Evidence Aggregation.}
To facilitate informed decision-making without relying on a single candidate model or checker outcome, we provide each selection agent with aggregate information across the synthesized semantic checkers, the subset of candidate models that passed execution checks G1--G3, and their semantic check outcomes summarized as a status overview for each candidate model (see Fig.~\ref{fig:CP-SynC} for an example). Each selection agent analyzes the aggregated information and uses it as evidence to independently critique all candidates against the problem context $p$, flagging inconsistencies or corner cases that might evade semantic checkers, as well as potential checker faults that could incorrectly reject a correct model.
\textbf{Stage 2: Final Model Selection.}
Conditioned on its own critiques, each agent votes to either select the model most aligned with the problem description or reject all candidates. The final decision is made by majority voting over the \(K\) agents.

\subsubsection{Staged Checking Pipeline}
\label{sec:checking_pipeline}
The checking pipeline is a staged cascade composed of three hard execution checkers (G1 syntax, G2 solver status, G3 output format) and a soft semantic checker (G4), surfacing targeted signals for repair. Failures from hard checkers must be cleared for a candidate model to proceed, while failures from the soft checkers do not enforce immediate rejection. Each checker is denoted as a function
\(
G_i(\mathcal{M}_{p}) = (\texttt{status}^{(i)},\, \texttt{feedback}^{(i)}),
\)
where \(\texttt{status}^{(i)} \in \{\texttt{pass},\texttt{fail}\}.\)
We denote by \(\mathcal{M}^{(k,t)}\) the candidate model produced by agent \(k\) after \(t\) refinement steps.
If $G_i$ reports a failure on $\mathcal{M}^{(k,t)}$, a repair is initiated to produce an updated candidate $\mathcal{M}^{(k,t+1)}$. Consequently, the checking process restarts, and the updated candidate re-enters the cascade at G1.
This rule prevents later checkers from evaluating models that may have drifted out of compliance with earlier checks due to a patch.

\begin{itemize}%[leftmargin=*, itemsep=0pt, parsep=0pt, topsep=0pt, partopsep=0pt]
    \item \textbf{G1: Syntax correctness.}  
    Compile the MiniZinc model with the MiniZinc CLI.
    \emph{Pass} if compilation succeeds.
    \emph{Fail} returns the compiler traceback.

    \item \textbf{G2: Solver status.}  
    Execute the model with a selected solver. \emph{Pass} if the solver reports a solution (e.g., \texttt{SATISFIABLE} or \texttt{OPTIMAL}). \emph{Fail} for \texttt{UNSAT}, \texttt{UNKNOWN}, or \texttt{TIMEOUT}, returning the solver status.
    
    \item \textbf{G3: Output format.} 
    Apply the formatter function $F_{p}$ to produce the output solution $\mathcal{S}_{p}$ and validate its conformity to $P_{OUT}$. \emph{Pass} if the format matches exactly. 
    \emph{Fail} returns a message detailing the discrepancies.

    \item \textbf{G4: Semantic check.}  
    Run the semantic checker suite $\{\mathcal{F}^{(k)}_{sc}\}_{k=1}^{K}$ on 
    $(\mathcal{I}_p, \mathcal{S}_{p})$.
    \emph{Pass} if a majority of checkers return \texttt{pass}.
    \emph{Fail} returns the set of failing checker IDs with feedback describing the violated requirements.

\end{itemize}

\section{Evaluation}
\subsection{Benchmark Suite}
We introduce an automated benchmark for comprehensive evaluation of our workflow.\footnote{\cmrd{The benchmark is available at \url{https://github.com/Yuliang795/LLMs-CP-CPSynC}.}} 
Each problem instance has the following items:
\begin{itemize}%[leftmargin=15pt, itemsep=0pt, parsep=0pt, topsep=0pt, partopsep=0pt]
    \item \textit{Problem description ($P_{NL}$):} natural-language problem requirements.
    \item \textit{Input specification ($P_{IN}$):} structure and meanings of the input data.
    \item \textit{Output format ($P_{OUT}$):} structure and meanings of the output solution.
    \item \textit{Input data (\(\mathcal{I}_{p}\)):} parameters provided in Python built-ins, and \texttt{.dzn} files.
    \item \textit{Reference model ($\mathcal{M}^{\mathrm{ref}}_p$):} a vetted CP formulation for solution validation.
    \item \textit{Solution-mapping function ($T_{p}$):} a function that converts an output solution \(\mathcal{S}_{p}\) into an assignment over reference variables.
\end{itemize}

\paragraph{Source and Curation}
Problems are drawn primarily from two well-established resources.
Problems 1--30 were adopted from \textsc{CPEval}~\cite{song2025llmcp} for the problem contexts $(P_{NL}, P_{IN}, P_{OUT})$ and input data.
Problems 31--100 are curated from \textsc{CSPLib} \cite{gent1999csplib} and the \textsc{PyCSP3} repository\footnote{https://github.com/xcsp3team/PyCSP3-models}, selected to ensure no overlap with the \textsc{CPEval} subset.
For each problem, we manually specified the output formats (\(P_{OUT}\)). 
Following prior work, we decoupled (\(P_{OUT}\)) from the reference formulation when applicable to avoid biasing generation toward specific formulations induced by the reference decision variables (See an example in Appendix~\ref{asec:eval}).
We then instantiated the corresponding solution-mapping function $T_{p}$ for each problem with an LLM-generated script, which we manually verified for correctness.
All problems are equipped with an evaluation script implemented following the protocol in \S\ref{sec:eval_protocol} \emph{Evaluation Protocol}. 
The resulting benchmark covers a broad range of domains, including industrial applications, academic problems, and puzzles.

\paragraph{Normalization}
We normalized each problem description by rectifying typos and removing extraneous content (e.g., hyperlinks, background notes, and ambiguous variants). When a problem description is incomplete or missing while a citation to the source literature is available, we extracted a faithful description from the cited source. We also reconciled discrepancies between the reference model and the problem description (e.g., removing symmetry-breaking constraints or constraints not implied by the requirements). Finally, for each problem, we required the reference model to solve the provided input instance within 30 seconds, and for COPs, the solution had to be optimal.

\subsection{Evaluation Protocol}
\label{sec:eval_protocol}
CP problems often admit multiple valid formulations with heterogeneous output formats \cite{frisch2005rules,apt2003principles}, making manual semantic alignment and validation challenging. Following prior work, we evaluate modeling performance at the solution level without inspecting the generated formulation \cite{michailidis2024constraint,song2025llmcp,michailidis2025cpbench,singirikonda2025text2zinc}. 
Specifically, for an output solution \(\mathcal{S}_p\), we first map it to an assignment over the reference model variables using the solution mapping function \(T_p\). We then check whether the assignment satisfies all constraints in the reference model for CSPs and COPs, and additionally verify optimality for COPs (See Appendix~\ref{asec:sol_val} for details).

\paragraph{Solution Accuracy}
For each problem instance $p \in \mathcal{P}$ with output solution $\mathcal{S}_p$, let $\mathrm{Feas}(\mathcal{S}_p)$ be a boolean indicating whether $\mathcal{S}_p$ satisfies all constraints of the reference formulation, and let $\mathrm{Opt}(\mathcal{S}_p)$ be a boolean indicating whether the optimal objective value is achieved (for COPs). We define the corresponding instance-level correctness indicator
$\gamma_p(\mathcal{S}_p) = \mathbf{1}[\mathrm{Feas}(\mathcal{S}_p)]$ if $p$ is a CSP, and $\mathbf{1}[\mathrm{Feas}(\mathcal{S}_p) \wedge \mathrm{Opt}(\mathcal{S}_p)]$ if $p$ is a COP.
\noindent The solution accuracy (SA) across the benchmark is then defined as:
\[
SA \;=\; \frac{1}{|\mathcal{P}|}\,\sum_{p\in\mathcal{P}} \gamma_p(\mathcal{S}_p).
\]

\subsection{Experimental Setup}
\label{sec:exp_setup}
\paragraph{LLM backbones}
We evaluate our method and the baselines on the proposed benchmark using several prominent LLMs, including 
GPT-4o (2024-08-06)\footnote{https://openai.com/index/gpt-4o-system-card/}, 
DeepSeek-V3 (0324) \cite{liu2024deepseekv3}, 
Claude-Sonnet-4 (20250514) \footnote{\url{https://www.anthropic.com/claude/sonnet}}, 
Qwen3-Next (Instruct)\footnote{\url{https://huggingface.co/Qwen/Qwen3-Next-80B-A3B-Instruct}}, and 
Gemini-2.5-flash-lite (preview-09-2025) \cite{comanici2025gemini}, hereafter denoted as Gemini.

\paragraph{\textsc{CP-SynC} configuration}
In the default setting ($K{=}3$, $r{=}4$, $R{=}1$), each iteration instantiates three agents per role to generate three candidate models, semantic checkers, and selection votes, allowing up to four self-refinements per candidate and one restart budget.
To induce diversity, for Qwen3-Next and Gemini, we use temperature sampling with \(\tau{=}0.7\).
For the remaining LLMs, we instantiate each agent role with three distinct problem descriptions (original, refined, and planning-augmented) following the diverse sampling
strategies in \S~\ref{sec:sampling_strategies}. 

\paragraph{Baselines}
We compare our approach to two recent workflows that support modeling in MiniZinc, namely the 2-Step (2S) method~\cite{song2025llmcp} and CP-Bench~\cite{michailidis2025cpbench}, and adopt the best-performing configurations reported in their literature. Detailed introduction and implementation details are provided in Appendix~\ref{asec:baseline_details}.

\paragraph{Modeling framework and solver}
All workflows use MiniZinc with the Gecode solver, subject to a 30-second timeout per solver call.

\section{Experimental Results}
\subsection{Modeling Performance}
Table~\ref{tab:all_methods} summarizes the performance of \textsc{CP-SynC} against two baselines, 2S and \textsc{CP-Bench}. 
Overall, \textsc{CP-SynC} consistently and often substantially outperforms both baselines across all tested LLMs. 
The improvements are particularly pronounced for backbones with lower initial SA. For example, comparing with the strong \textsc{CP-Bench} baseline, $\textsc{CP-SynC}$ boosts DeepSeek-V3's performance from $56.7 \rightarrow 77.7$ and GPT-4o from $65.3 \rightarrow 78.0$. 
The smallest model tested, Gemini-2.5-flash-lite, exhibited the most dramatic improvement, rising from $19.0$ (2S) and $48.3$ (\textsc{CP-Bench}) to $67.0$ with \textsc{CP-SynC}, and it also surpasses baselines with several larger backbones.
Among all evaluated backbones, Claude-Sonnet-4 remains the strongest overall, exhibiting robust MiniZinc fluency and strong intrinsic modeling ability (see Appendix~\ref{asec:error_type_analysis} for error type analysis). The incremental gain of \textsc{CP-SynC} is modest when the backbone LLM already yields high-quality initial candidates.

\begin{table}[!h]
    \caption{Benchmark scores (\textbf{SA \%}) across modeling systems and backbones. Each method is run three times and we report the average. The best-performing method for each LLM is underlined and the best overall score is bolded.
    ``Sampling'' indicates the diverse sampling strategy used by \textsc{CP-SynC} for each LLM backbone (P: diverse prompt sampling; T: temperature sampling).}
    \label{tab:all_methods}
    \centering
    \small
    \setlength{\tabcolsep}{3.5pt}
    \begin{tabular}{lcccc}
        \toprule
        \textbf{Model} & \textbf{Sampling} &\textbf{2S} & \textbf{CP-Bench} & \textbf{CP-SynC} \\
        \midrule
        DeepSeek-V3 & P &54.0 & 56.7 & \underline{77.7} \\
        GPT-4o & P & 54.3 & 65.3 & \underline{78.0} \\
        Claude-Sonnet-4 & P & 74.7 & 84.3 & \underline{\textbf{86.3}} \\
        Qwen3-Next & T & 59.7 & 68.0 & \underline{82.3} \\
        Gemini 2.5-flash-lite & T & 19.0 & 48.3 & \underline{67.0} \\
        \bottomrule
        \end{tabular}%
\end{table}

We attribute these gains to \textsc{CP-SynC}'s design of test-driven multi-trajectory modeling.
The workflow instantiates diverse initial modeling trajectories, mitigating the effects of path fixation on an early erroneous modeling choice and increasing the likelihood of converging to a correct formulation 
(see \S\ref{sec:scale_iter} for detailed analysis). 
The multi-agent collaboration enables test-driven development via synthesized semantic checkers that provide immediate feedback on semantic correctness for the generated models.

\subsection{Ablation Study}
\label{sec:ablation_study}

\begin{table}[ht!]
\caption{Ablation over subsets of agent types. All configurations are run three times, and the average is reported. ``Gemini'' refers to \emph{Gemini 2.5-flash-lite}.}
\label{tab:config_ablation}
\centering
\small
\setlength{\tabcolsep}{3pt}
\begin{threeparttable}
\begin{tabular}{cccc|cc|cc}
\toprule
\multirow{2}{*}{\textsc{Config.}} & \multirow{2}{*}{\textsc{Modeling}} & \multirow{2}{*}{\textsc{Validation}} & \multirow{2}{*}{\textsc{Selection}} & \multicolumn{2}{c|}{\textbf{Qwen3-Next}} & \multicolumn{2}{c}{\textbf{Gemini}} \\
\cmidrule(lr){5-6} \cmidrule(lr){7-8}
 & & & & \textbf{SA} & \textbf{SA@1} & \textbf{SA} & \textbf{SA@1} \\
\midrule
1a & \checkmark & & & 47.7 & -- & 22 & -- \\
1b & \checkmark & & & 69 & -- & 41.3 & -- \\
\midrule
1 & \checkmark & & & 78.0 & 86.3 & 64.3 & 70.0 \\
2 & \checkmark & \checkmark & & 77.7 & 83.7 & 63.7 & 66.0 \\
3 & \checkmark & & \checkmark & 78.7 & 84.3 & 65.0 & 67.0 \\
4 & \checkmark & \checkmark & \checkmark & 82.3 & 83.0 & 67.0 & 68.3 \\
\bottomrule
\end{tabular}
\begin{tablenotes}[flushleft]
\footnotesize
\item 1a: Single-trajectory modeling without self-refinement.
\item 1b: Single-trajectory modeling with up to four self-refinements.
\end{tablenotes}
\end{threeparttable}
\end{table}

To isolate the contribution of each module, we decompose \textsc{CP-SynC} into distinct agent subsets by enabling or disabling certain agent types.
We also include two single-trajectory variants of the modeling agent that generate a single model with and without self-refinement ($r{\in}\{0,4\}$).
Full configurations are in Appendix~\ref{asec:ablation_study}. Due to API cost, this study is conducted using Qwen3-Next and Gemini. 

We report SA and compute the at-least-one pass rate (SA@1), defined as the proportion of problems where at least one sampled model passes evaluation, representing the oracle upper bound given the candidate pool. 
The oracle gap SA@1$-$SA indicates how often correct models are sampled but not selected.

Table~\ref{tab:config_ablation} highlights three main takeaways. 
First, 
self-refinement is highly beneficial, substantially improving SA for both LLMs in the single-trajectory setting (Config. 1a vs. 1b). 
Second, 
multi-trajectory modeling produces a strong candidate pool, but the ablated variants that rely on a single source of evidence for selection (e.g., Config.~1, majority voting over output solutions) leave a noticeable gap to the oracle SA@1, indicating that correct candidates are often generated but not reliably selected. 
Third, 
naively refining the generated model against synthesized checkers (Config 2) \cmrd{underperforms the other multi-trajectory variants}, which our inspection attributes to strict refinement against noisy semantic checker feedback 
(See Appendix~\ref{asec:syn_checker} for an empirical study quantifying the adverse effects of synthesized checkers).
In contrast, the full pipeline achieves the best SA overall and nearly closes the oracle gap. We provide a detailed analysis in Appendix~\ref{asec:ablation_study}.

\subsection{Test-Time Scaling Strategies}
\label{sec:scale_iter}
\begin{figure}[ht!]
    \centering
    \includesvg[width=\textwidth]{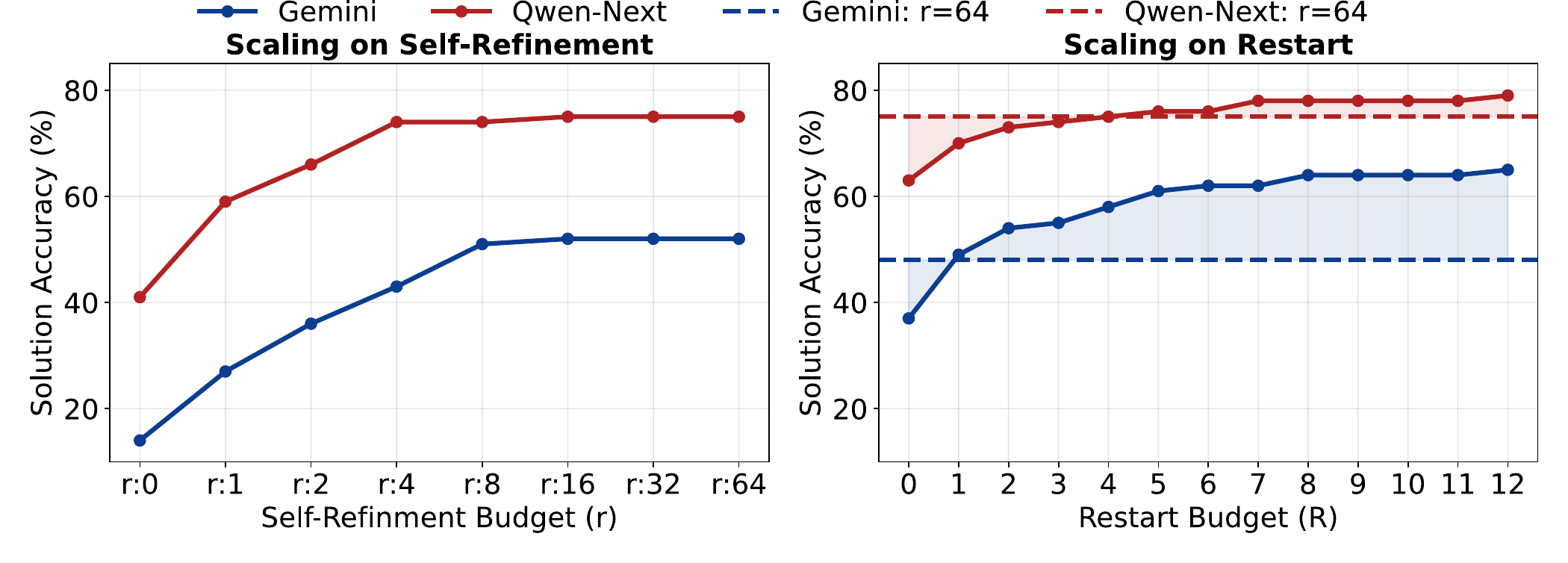}
    \caption{Impact of budget allocation on SA score: (left) single-trajectory generation with an increasing self-refinement budget; (right) multi-trajectory generation via restarts with an increasing restart budget and fixed self-refinement budgets.}
    \label{fig:iter_scaling}
\end{figure}

Our ablation study shows that LLMs often fail to generate valid CP models in a single shot without self-refinement.
However, solely refining along a single trajectory can underperform due to refinement inertia: once an LLM produces an erroneous candidate, it may persist in reproducing the same error during subsequent self-refinements~\cite{huang2023llmstubborn}. 
This motivates finding a balance between refining an erroneous candidate to exploit a promising but potentially flawed trajectory and starting a new modeling trajectory to escape local minima.

In this experiment, we use CP-SynC as the backbone workflow and study the impact on SA when allocating a total budget of 64 LLM calls to either refining an erroneous candidate against the staged checking pipeline or starting a new modeling trajectory. The left panel of Fig.~\ref{fig:iter_scaling} illustrates single-trajectory generation with an increasing self-refinement budget ($r$). For Gemini, SA is lowest at $r=0$ and improves rapidly as $r$ increases to 8, after which SA plateaus at approximately 52\%. 
A similar trend is observed for Qwen3-Next, where SA rises quickly and converges by $r=4$ at around 76\%. 
The right panel depicts multi-trajectory generation via restarts under the same call budget. Here, we cap self-refinement at $r=4$ iterations per modeling trajectory to match our default configuration in \S~\ref{sec:exp_setup}, and start a new trajectory if this limit is exceeded without producing a valid model, allowing up to $R$ restarts. For a fair comparison, we cap the restart budget at \(R=12\) to match the same total budget of 64 LLM calls used in the left panel.\footnote{Excluding the initial generation call, we enforce $r + (1{+}r)R \le 64$, yielding $R{=}12$.} The dashed line indicates the best SA achieved by solely refining a single modeling trajectory (left panel). Notably, for Gemini, multi-trajectory generation already surpasses this baseline at $R=2$. As $R$ increases, SA rises rapidly and then plateaus around \(66\%\) at \(R=8\). For Qwen3-Next, multi-trajectory generation achieves comparable SA to the baseline at $R=4$ and surpasses it thereafter, although the gains are more limited.

\section{Related Works}
Recent work explores LLMs for several roles in constraint programming, including constraint modeling, end-to-end solving~\cite{michailidis2024constraint,jiang2025llmsolve}, and model refinement \cite{voboril2025streamliner}. Our work focuses on automated constraint modeling. In this direction, prior work commonly relies on three recurring components: (i) few-shot prompting, where solved exemplars are retrieved at inference time as modeling hints~\cite{michailidis2024constraint,michailidis2025cpbench,singirikonda2025text2zinc,szeider2025cpagent}; (ii) intermediate representations, such as formulations distilled from the problem statement~\cite{michailidis2024constraint} or constraint models expressed as knowledge graphs to expose modeling logic~\cite{singirikonda2025text2zinc}; and (iii) domain-guided prompting, where curated modeling strategies are incorporated into the prompt to steer model construction~\cite{szeider2025cpagent,michailidis2025cpbench}.
Among closely related systems, \textsc{CP-Agent} is an agentic framework for modeling in CPMPy. It adopts a refinement loop with code execution and LLM-synthesized model validation functions to certify the solutions before yielding final models~\cite{szeider2025cpagent}.\footnote{We exclude \textsc{CP-Agent} from our baselines as it does not currently support MiniZinc.}
\textsc{CP-Bench} applies test-time scaling via repeated sampling, solution-level majority voting, and iterative self-refinement guided by LLM critiques~\cite{michailidis2025cpbench}.

\section{Conclusion and Discussion}
This work presents \textsc{CP-SynC}, a novel multi-agent workflow for zero-shot constraint modeling in MiniZinc. 
\textsc{CP-SynC} coordinates specialized agents for modeling, validation, and selection. To mitigate the noise inherent in LLM generation, we employ diverse sampling strategies within each role and explore multiple modeling trajectories in parallel, increasing the likelihood of discovering a semantically correct model.
Addressing the absence of oracle validation at test time, we utilize synthesized semantic checkers as approximate supervision to guide model refinement and selection. 
Experiments on a benchmark of 100 CP problems from diverse domains show that \textsc{CP-SynC} achieves \cmrd{the strongest MiniZinc modeling performance among the evaluated baselines}.
Our ablation study quantifies the \cmrd{synergy among the roles within the proposed agentic} architecture and provides practical insights for test-time scaling under budget constraints.

Our work highlights several interesting directions for future research. First, our results on multi-trajectory modeling suggest a favorable balance between exploring new trajectories and refining erroneous candidates. However, we adopt a fixed budget allocation strategy. Future work could develop heuristics or learned policies that dynamically allocate the inference budget.
Second, training specialized small language models offers a promising path toward more efficient and accessible automation of constraint modeling. Third, systematic user studies are needed to understand how domain experts interact with LLM-based modeling systems, informing the design of more usable future systems.

\bibliographystyle{splncs04}
\bibliography{samplepaper}

\clearpage
\appendix
\section{Illustrative Examples}
\subsection{N-Queens Example}
\label{asec:eg_nqueens}
In this section, we use the classic N-Queens problem to demonstrate how MiniZinc encodes a problem description into a machine readable model.
We show two equivalent formulations: Model~(a) explicitly enforces pairwise inequalities, while Model~(b) uses the global constraint \texttt{all\_different}. The \texttt{all\_different} constraint enforces pairwise inequality among a set of variables and can replace a quadratic number of explicit \(\neq\) constraints.

\paragraph{Problem Description} Place $n$ queens on an $n\times n$ grid such that no two queens share the same row, column, or diagonal.

\medskip
\begin{tcolorbox}[promptbox,title={(a) MiniZinc model without global constraints}]
\begin{verbatim}
int: n;
array[1..n] of var 1..n: q;

constraint
  forall(i, j in 1..n where i < j) (
    q[i] != q[j] /\
    q[i] + i != q[j] + j /\
    q[i] - i != q[j] - j
  );

solve satisfy;
\end{verbatim}
\end{tcolorbox}

\medskip
\begin{tcolorbox}[promptbox,title={(b) MiniZinc model with global constraints}]
\begin{verbatim}
int: n;
array[1..n] of var 1..n: q;

constraint
  all_different(q) /\
  all_different([ q[i] + i | i in 1..n ]) /\
  all_different([ q[i] - i | i in 1..n ]);

solve satisfy;
\end{verbatim}
\end{tcolorbox}

\section{Experimental Settings}
\label{sec:appendix}
\subsection{Evaluation}
\label{asec:eval}

\subsubsection{Solution mapping}
\label{sec:sol_mapping}
For each problem \(p\), we specify a subset of decision variables \(X^{\mathrm{map}}_{p} \subseteq X^{\mathrm{ref}}_{p}\) for verification, and a problem-specific solution-mapping function \(T_{p}:\ \mathcal{S}_{p} \to \prod_{x\in X^{\mathrm{map}}_{p}} D^{\mathrm{ref}}_{p}(x)\), which converts the output solution \(\mathcal{S}_{p}\) into an assignment \(\tilde{x} = T_p(\mathcal{S}_{p})\) over \(X^{\mathrm{map}}_{p}\).

\emph{Example (N-Queens).} Consider a reference formulation that uses an array $q[1..n]$, where $q_i$ denotes the column of the queen in row $i$. Enforcing this specific output format may bias generation towards a narrow family of formulations that adopt this encoding, and it also artificially simplifies the task, since the variable definition itself implicitly prevents row collisions.
As a more canonical task level format, one could instead use a binary matrix $B \in \{0,1\}^{n \times n}$ with $B_{i,j} = 1$ iff a queen is placed at $(i,j)$. For evaluation, we accompany this with a transformation function $T_{p}$ that converts $B$ into the reference encoding $q$ via
\(
q_i = \sum_{j=1}^n j \cdot B_{i,j}
\).

\subsubsection{Solution Validation}
\label{asec:sol_val}
We assess the correctness of a generated output \(\mathcal{S}_{p}\) by checking that it satisfies all constraints in the reference model for both CSPs and COPs, and additionally testing solution optimality for COPs.

\paragraph{CSP validation}
For CSPs, we inject the mapped assignment $\tilde{x}$ into the reference model via equality constraints:
\(
\mathrm{Eq}(\tilde{x}) = \{ x = \tilde{x}(x)\ :\ x \in X^{\mathrm{map}}_p \}.
\)
A solution $\mathcal{S}_{p}$ is valid iff the augmented reference model is satisfiable:
\begin{equation}
Feas(\mathcal{S}_{{p}}) = \mathrm{SAT}\Bigl(X^{\mathrm{ref}}_p,\ D^{\mathrm{ref}}_p,\ C^{\mathrm{ref}}_p \cup \mathrm{Eq}(\tilde{x})\Bigr).
\label{check:feas}
\end{equation}

\paragraph{COP validation}
We first check feasibility as in Eq.~\eqref{check:feas}, requiring that the mapped assignment satisfies all constraints and that the reported objective value matches what the reference model computes for that assignment.
Next, we test optimality by asking whether the reference model admits a strictly better objective value than the reported one.

Let $z$ denote the objective value corresponding to $\mathcal{S}_{p}$, we solve the reference model with an augment dominance constraint to check for a superior solution:
\begin{equation}
Opt(\mathcal{S}_{p}) = 
\begin{cases} 
\neg \mathrm{SAT}\Bigl(X^{\mathrm{ref}}_p, D^{\mathrm{ref}}_p, C^{\mathrm{ref}}_p \cup \{ f^{\mathrm{ref}}_p(x) < z \}\Bigr) & \text{minimization}, \\[6pt]
\neg \mathrm{SAT}\Bigl(X^{\mathrm{ref}}_p, D^{\mathrm{ref}}_p, C^{\mathrm{ref}}_p \cup \{ f^{\mathrm{ref}}_p(x) > z \}\Bigr) & \text{maximization}.
\end{cases}
\label{check:opt}
\end{equation}

\noindent If this check yields \texttt{True} (i.e., the instance is \texttt{UNSATISFIABLE}), no strictly better solution exists and the candidate is optimal; otherwise, it is suboptimal.

\subsection{Benchmark Validation}
\label{asec:benchmark_val}
To assess the reliability of the evaluation protocol, we cross-reference results against the oracle unit tests provided in \textsc{CPEval}~\cite{song2025llmcp} on the subset of overlapping problems.
Using the Two-Step (2S) method, we sampled candidate models with representative LLMs (GPT-4o and DeepSeek-V3) and evaluated them under both protocols to assess concordance.
Specifically, we define the mutual agreement score as the proportion of models that receive identical verdicts (pass or fail) under both protocols. 
Empirical results demonstrate near-perfect mutual agreement between the proposed benchmark and the human-crafted unit tests of \textsc{CPEval}, confirming that our automated protocol provides a reliable and scalable alternative to manual expert validation.

\subsection{Implementation Details}
\subsubsection{Baselines} For a fair comparison, we include recent open-source workflows that support modeling in MiniZinc:
\label{asec:baseline_details}
\begin{itemize}[leftmargin=*, itemsep=0pt, parsep=0pt, topsep=0pt, partopsep=0pt]
    \item \textbf{Two-Step (2S):} The best performing workflow from \textsc{CPEVAL}~\cite{song2025llmcp} that (i) generates a CP model and refines it against execution feedback and (ii) emits the solutions in the required formats. We run the official configuration, sampling one candidate per problem at temperature $0$ and up to three self-refinement attempts.

    \item \textbf{CP-Bench:} The workflow from \textsc{CP-Bench}~\cite{michailidis2025cpbench} employs expert-curated CP modeling instruction prompts, repeated model sampling, and self-verification. We use their best configuration reported for MiniZinc and apply their best-performing test-time scaling configuration, corresponding to 10 sampled models, 10 self-verification attempts, and a level-3 system prompt.
\end{itemize}

\section{Experimental Results}
\subsection{Efficacy of Self-Checking Mechanisms in CP} 
\label{asec:syn_checker}
\begin{table}[b!]
\caption{False rejection rate (FRR, lower is better) for two self-checking signals. ``Syn. Checker'' denotes LLM-synthesized semantic checkers, and ``Critique'' denotes LLM-generated critiques.}
\label{tab:checker_eval}
\setlength{\tabcolsep}{4pt}
\centering
\begin{tabular}{lcc}
\toprule
Model & Syn. Checker & Critique \\
\midrule
DeepSeek-V3           & 0.227 & \textbf{0.283} \\
GPT-4o                & 0.243 & 0.360 \\
Claude-Sonnet-4     & \textbf{0.147} & 0.463 \\
Qwen3-Next            & 0.253 & 0.730 \\
Gemini & 0.260 & 0.720 \\
\bottomrule
\end{tabular}
\end{table}

Self-checking mechanisms are commonly adopted in LLM-based modeling workflows, serving as proxies for semantic correctness to guide model refinement. However, when their feedback is flawed, naively refining against it can steer generation toward semantic misalignment. In this section, we conduct empirical experiments to quantify these adverse effects by measuring how often self-checking feedback incorrectly identifies flaws in a correct artifact. Our analysis focuses on two popular self-checking mechanisms: LLM-synthesized checkers and critiques. For checkers, we prompt each LLM to synthesize checker functions from the problem description using the same prompts as the validation agents in \S\ref{sec:imp_detail}, and then apply these checkers to check the reference solutions provided by the benchmark. For LLM-generated critiques, we prompt each LLM to review the reference model and reason whether it aligns with the problem description. We classify any checker or critique response that incorrectly identifies a flaw in a valid ground-truth component as a false rejection. Specifically, a false rejection occurs when (i) a synthesized checker rejects the reference solution, or (ii) a generated critique rejects the reference model.
Correspondingly, we define the false rejection rate (FRR) as the fraction of ground-truth components incorrectly rejected across all problems.
To evaluate this, we sample three checkers and three critiques per LLM for each problem and report the average FRR with a temperature of \(\tau=0\).

Table~\ref{tab:checker_eval} shows that both self-checking signals exhibit non-negligible FRR across all tested LLMs. For synthesized checkers, Claude-sonnet-4 achieves the lowest FRR (14.7\%) followed by DeepSeek-V3 (22.7\%), while Gemini exhibits the highest (26\%). The LLM-generated critiques are substantially less reliable, with FRR ranging from 28.3\% (DeepSeek-V3) to over 73\% (Qwen3-Next). These results suggest that naively refining against noisy LLM-based self-checking can steer generation away from true semantic alignment, and that successful modeling requires not only effective refinement with self-checking, but also the ability to recognize faulty self-checking signals.

\subsection{Ablation Study}
\label{asec:ablation_study}
\paragraph{Implementation Details}
To study the effect of each module, we decompose the multi-agent architecture into ablated configurations by enabling or disabling subsets of agents:
(1) modeling agents only, which samples model candidates, performs self-refinement with checkers G1--G3, and then chooses the final model by solution level majority. To quantify the effect of our diverse sampling and refinement techniques, we also include two single-trajectory modeling agents baselines: generating a single model without self-refinement \((r{=}0)\) and generating a single model with up to four self-refinements \((r{=}4)\). (2) Modeling Agents + Selection Agents, which is identical to (1) but replaces solution-majority with a multi-agent, consensus-based selector. (3) Modeling Agents + Validation Agents, which adds synthesized semantic checkers (G4) as approximate verification signals within the self-refinement loop and selects the candidate that passes the most semantic checkers. (4) Full pipeline, which activates all three modules. Ties are broken deterministically by first occurrence, and all other settings remain constant.

\paragraph{Detailed Analysis}
The single-trajectory baselines highlight the benefit of self-refinements, as increasing the refinement depth from $r{=}0$ to $r{=}4$ markedly improves SA for both LLMs
(Qwen3-Next: 47.7 $\to$ 69.0; Gemini: 22.0 $\to$ 41.3), confirming that a few refinement steps can substantially rescue an initial model from a single erroneous trajectory.
Moving to the multi-candidate configurations, the Modeling Agents only configuration already produces a strong pool for both LLMs, but leaves a sizable gap to the oracle (Qwen3-Next: SA=78 vs SA@1 = 86.3; Gemini: SA=64.3 vs SA@1=70), indicating that correct solutions are present but are not reliably selected by solution level majority voting. 
Replacing solution majority selection with the consensus-based Selection Agents (Config 3) yields marginally higher SA. 
Introducing the Validation Agents (Config 2) slightly narrows the SA@1 (83.7 for Qwen3-Next and 66.0 for Gemini) but slightly reduces SA compared to Config 1, which our inspection attributes to occasional noise.
In contrast, the full pipeline (Config 4) attains the best SA for both models (82.3 for Qwen3-Next, 67.0 for Gemini) while nearly closing the oracle gap (SA@1 = 83.0 and 68.3, respectively). In this setting the selection agents can jointly consider candidate code and feedback from semantic checks, enabling them to down-weight suspected checker noise and make more informed choices.

\subsection{Error-Type Analysis}
\label{asec:error_type_analysis}
\begin{figure}[ht!]
    \centering
    \includesvg[width=\textwidth]{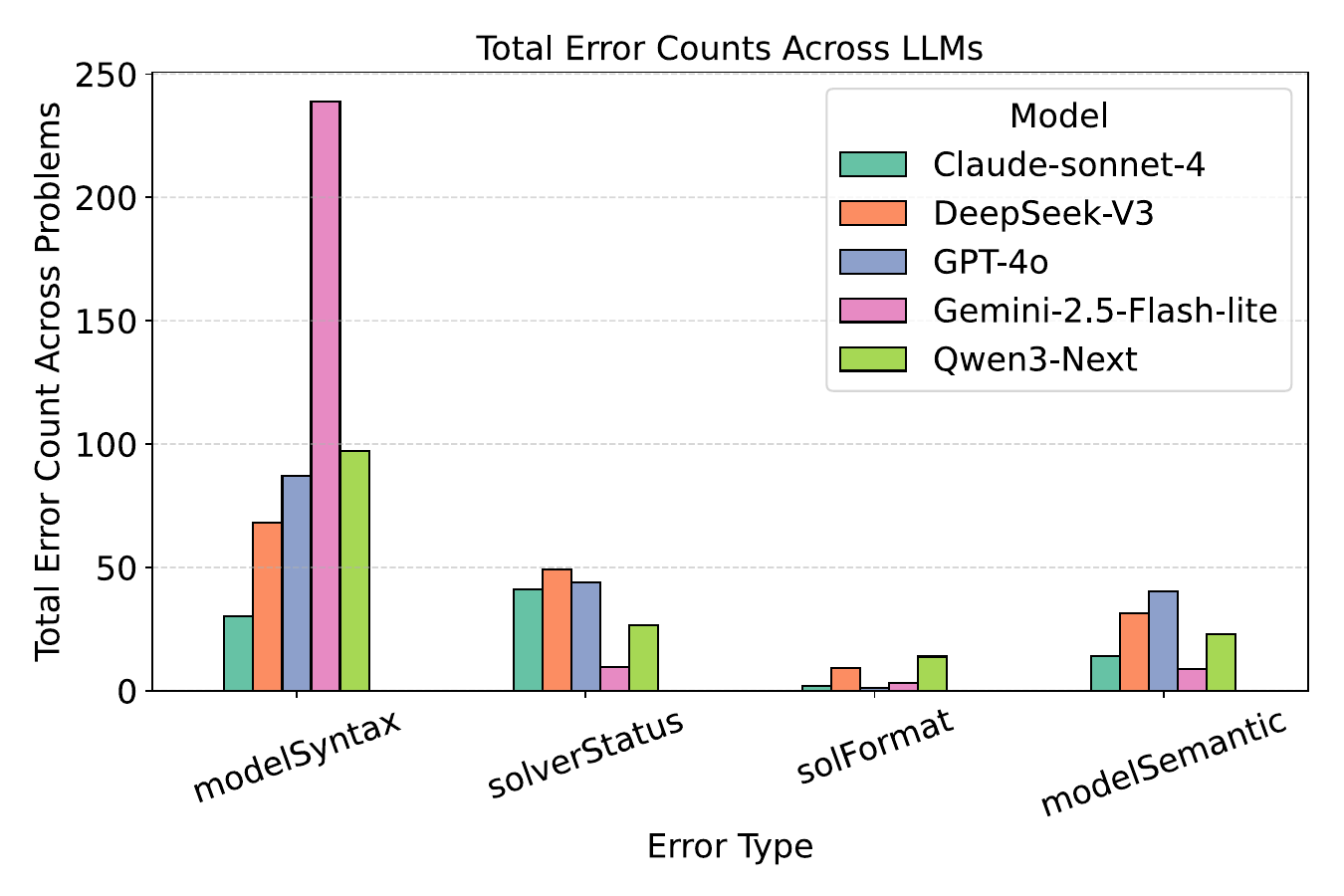}
    \caption{Error counts by checks (G1--G4) across LLMs.}
    \label{fig:error_comparison}
\end{figure}

We categorize modeling failures according to the four checkers from the staged checking pipeline G1--G4 (Section~\ref{sec:checking_pipeline}). 
Figure~\ref{fig:error_comparison} presents the histogram of total error counts across all LLMs.
Overall, syntax error (G1) is the dominant failure mode across LLMs, suggesting a comparatively weaker fluency in MiniZinc syntax. Conversely, solution formatting errors (G3) are least frequent, indicating that tested LLMs can reliably adhere to the prescribed output format once a valid solver output is produced. Semantic failures related to the solver (G2) and semantic checkers (G4) are moderate. 
Among the tested LLMs, Claude-Sonnet-4 exhibits the fewest total errors, especially semantic errors, showing high internal consistency across modeling, testing, and selection.
Interestingly, Gemini produces the highest number of syntax errors while showing few solver-status and semantic errors, likely because repeated syntax issues exhaust the refinement budget and prevent progression to later checkers.

\section{Prompts}
\label{asec:prompts}
\subsection{Agents System Prompts}
This section lists the system prompts used for each agent role. For modeling agent, we include the system prompt for model generation and output formatting here, and present the self-refinement prompts in Appendix~\ref{asec:prompt:self-refinement}.

\begin{tcolorbox}[promptbox,title={System Prompt: Modeling Agent (Modeling)}]
\footnotesize
\begin{Verbatim}[
  breaklines=true,
  breakanywhere=true,
  breaksymbolleft=,
  breaksymbolright=
]
You are an expert in constraint programming. Your task is to develop a MiniZinc model based on a provided problem description. You will be given: 
(1) Problem description - read and analyze the problem carefully.
(2) Input parameter specification - Parameters will be provided via a .dzn file during execution manually. Do not generate or include any example .dzn content.
(3) Required output formats - The solution output will be formed based on the solution of the model in a later stage; for this task, do not include any code for displaying the solution or decision variable values in the model.

For example, for the N-Queens problem: 
- problem description:
"Can n queens be placed on a n by n chessboard so that no two Queens are on the same row, column or diagonal"
- details of the parameters:
"n": "The size of the chessboard and the number of queens, given as an integer."
- required output formats
(1) `queens`: "An array representing the positions of the queens on the chessboard. The value at each index `i` represents the row position of the queen in the ( i )-th column.","size": "[n]"

Given the problem description, the main semantic constraints are identified as no row, column, or diagonal conflicts. Therefore, develop the corresponding minizinc code:
```MiniZinc
% include needed libraries
include "globals.mzn";
% load parameters
int: n;
% instantiate decision variables, each queen is in a different row
array[1..n] of var 1..n: queens;
% no two queens in the same column
constraint all_different(queens);
% no two queens on the same diagonal
constraint
    forall(i, j in 1..n where i < j) (
         queens[i] + i != queens[j] + j /\
         queens[i] - i != queens[j] - j
    );
% solve the problem
solve satisfy;
```

Note that you must carefully analyze the problem description and determine the domain of each decision variable, also reason about the indexing for arrays(0-based or 1-based), as well as the constraints that need to be applied. Do not assume any specific constraints or domains unless they are explicitly stated in the problem description.

Now, based on the following problem context, write the corresponding MiniZinc model:
Problem description:
{Problem description}

Input parameters:
{Specification of the input parameters}

Required output formats:
(Reminder: Do not include any code for displaying or printing outputs.)
{Specification of the required output formats}
\end{Verbatim}
\end{tcolorbox}

\begin{tcolorbox}[promptbox,title={System Prompt: Modeling Agent (Output Formatting)}]
\footnotesize
\begin{Verbatim}[
  breaklines=true,
  breakanywhere=true,
  breaksymbolleft=,
  breaksymbolright=
]
The MiniZinc model has been successfully executed, and the decision variable values have been extracted into a Python dictionary using standard Python data types: {dvar_info}. Your task is to write a Python function called 'transformer' that transforms the decision variable values into a the specified format, and returns them in a dictionary, using keys as specified in the requirements. Carefully analyze how each decision variable is defined and structured in the generated model, and at the beginning of your function, add a comment that briefly lists all available decision variables and gives a one-sentence description for each. Do not include any test code or extraneous output - only the `transformer` function.

This function should take two parameters: (1) data_dict, which contains all the input parameter values, and (2) decision_var_dict, which contains exactly the decision variable values as used in the MiniZinc model (already converted to Python built-in types such as integers, arrays, strings, etc.).

The transformation logic may involve complex calculations; however, use the decision variable values directly if they already match the required output format. Ensure your Python script handles these transformations accurately and returns a dictionary with variables matching the following format:
{Specification of the required output formats}
\end{Verbatim}
\end{tcolorbox}

\begin{tcolorbox}[promptbox,title={System Prompt: Validation Agent}]
\footnotesize
\begin{Verbatim}[
  breaklines=true,
  breakanywhere=true,
  breaksymbolleft=,
  breaksymbolright=
]
You are an constraint programming modeling expert. Generate a Python function that performs semantic validation for the solution of a CP problem. The function must validate all semantic constraints implied by the problem description explicitly and precisely. For each type of semantic or structural violation, raise a ValueError with a clear, specific failure reason. This function should take two parameters: (1) data_dict, which contains all the input parameter values, and (2) ovar_dict, which contains output solutions in the predefined formats.
For example, for the N-Queens problem: 
- problem description:
"Can n queens be placed on a n by n chessboard so that no two Queens are on the same row, column or diagonal."
- details of the data_dict:
"n": "The size of the chessboard and the number of queens, given as an integer."
- output solutions and formats
(1) `queens`: An array representing the positions of the queens on the chessboard. The value at each index `i` represents the row position of the queen in the ( i )-th column.","size": "[n]"

Given the problem description, the main semantic constraints are extracted as no row, column, or diagonal conflicts. Therefore, develop the checker function below to check for row and diagonal conflicts between queens and raise specific errors such as: raise ValueError("Error: Queens at column 2 and 4 are in the same row.")
```Python
def semantic_checker(data_dict, decision_var_dict):
    n = data_dict["n"]
    queens = decision_var_dict["queens"]
    for i in range(n):
        for j in range(i + 1, n):
            if queens[i] == queens[j]:
                raise ValueError(f"Row conflict: queens at column {{i + 1}} and {{j + 1}} are both in row {{queens[i]}}")
            if abs(queens[i] - queens[j]) == abs(i - j):
                raise ValueError(f"Diagonal conflict: queens at column {{i + 1}} and {{j + 1}} are on the same diagonal")
```
Do not include any test code or extraneous output - only the `semantic_checker` function.

The following are the context of the current problem:
Problem description:
{Problem description}

Details of the data_dict:
{Specification of the input parameters}

Available output solutions in the output_dict:
{A list of decision variables used in the generated model}
\end{Verbatim}
\end{tcolorbox}

\begin{tcolorbox}[promptbox,title={System Prompt: Selection Agent}]
\footnotesize
\begin{Verbatim}[
  breaklines=true,
  breakanywhere=true,
  breaksymbolleft=,
  breaksymbolright=
]
You are an expert in constraint programming and MiniZinc modeling. For a given problem, you will be given multiple code candidates, each wrapped in the following format:
<candidate i>
```MiniZinc
...code of the MiniZinc model...
```
</candidate i>

All code candidates are syntactically correct and have been solved to a solution. Your task is to:
1. Carefully inspect each candidate for semantical correctness, identifying any issues in their logic, modeling approach, or constraint formulation. i.e.:
  - check if each decision variable is defined correctly with reasonable domain and indexing if applicable.
  - check if the constraints are correctly formulated align with the problem description.
  - ignore the output solution formats, as they will be converted to the required format in a later stage.
2. Review the checker results for each candidate, and consider any reported issues in your evaluation.
  - Note that the test checkers were synthesized and may contain flaws or syntax errors. Carefully review the checkers' logic, you should reject and ignore the feedback if: (1) you believe the code candidates are correct and the failures are due to flaws in the checker's logic, or (2) the error is caused by a defect in the checker itself (e.g., a syntax error).
  - Do not assume constraints not explicitly stated in the problem description, but appeared in the checkers.
3. Review the output solutions for each candidate, and consider any discrepancies with the problem requirements in your evaluation.
4. Select the candidate aligns with the problem description most overall. If all candidates are flawed, state the reason and use index -1 as the selection.
  - if multiple candidates are correct, select the one with the most precise, complete, and efficient modeling.

Now review the following code candidate(s):
Problem description:
{Problem description}

Synthesized semantic checkers:
{Code of all semantic checkers}

Candidate models:
{Code of all candidate models}

Among the candidates:
{Checker outcomes for each candidate}


Return your answer **only** as JSON in the following format:  
{{
  "reason": "<Concise reasoning of the defects in unchosen model (less than 5 sentences) and why the chosen candidate is the best.>",
  "selection": i
}}
\end{Verbatim}
\end{tcolorbox}

\subsection{Self-Refinement}
\label{asec:prompt:self-refinement}
\begin{tcolorbox}[promptbox,title={Prompt for repairing syntax errors}]
\footnotesize
\begin{Verbatim}[
  breaklines=true,
  breakanywhere=true,
  breaksymbolleft=,
  breaksymbolright=
]
The code you generated caused a runtime error with the following message:
```
{error message}
```

Please review the error message and the code, explain why this error occured in one sentence, then fix the error and provide the correct code in a code block.
Reference the related usage from documentations, check carefully if you are using the correct syntax.
Do not include any code for displaying or printing the solution or decision variable values in any format. The model will be evaluated manually.

\end{Verbatim}
\end{tcolorbox}

\begin{tcolorbox}[promptbox,title={Prompt for repairing solver status failures}]
\footnotesize
\begin{Verbatim}[
  breaklines=true,
  breakanywhere=true,
  breaksymbolleft=,
  breaksymbolright=
]
The MiniZinc model you generated was executed and resulted in an unsatisfiable result. Please carefully review the model and explain the likely cause of the failure in one sentence. Then regenerate the corrected MiniZinc code in a markdown code block. Assume that all input parameters are valid and within reasonable bounds. Do not omit any parameters or required constraints based on the assumption that the model is too restrictive. Your correction should preserve the full intent of the original problem description.
{Code of the current model}

\end{Verbatim}
\end{tcolorbox}

\begin{tcolorbox}[promptbox,title={Prompt for repairing output format errors}]
\footnotesize
\begin{Verbatim}[
  breaklines=true,
  breakanywhere=true,
  breaksymbolleft=,
  breaksymbolright=
]
The 'transformer' function was executed, and its output was passed to a format-checking function, which returned the following error: 
```
{error message}
```
Please review the error message regarding the solution format, explain the cause of the error in one sentence, and then fix the error and provide the correct 'transformer' function code in a markdown code block. 

\end{Verbatim}
\end{tcolorbox}

\begin{tcolorbox}[promptbox,title={Prompt for refining based on semantic checker feedback}]
\footnotesize
\begin{Verbatim}[
  breaklines=true,
  breakanywhere=true,
  breaksymbolleft=,
  breaksymbolright=
]
The code you generated was evaluated using multiple unit tests. Below is a summary of the test outcomes:
```
{Feedback from all semantic checkers for the current model}
```

Note that the test checkers were synthesized and may contain flaws or syntax errors. Carefully review the checkers' logic, you should only provide revised code if the test failures clearly indicate a problem in your code. You should reject the feedback if: (1) you believe your code is correct and the failures are due to flaws in the checker's logic, or (2) the error is caused by a defect in the checker itself (e.g., a syntax error).
Return your answer in the following JSON format:
If refinement is needed:
{{
  "reason": "<One sentence explanation based on the test feedback for each checker>",
  "decision": "accept",
  "revised_code": "<Revised code>"
}}
If refinement is not needed:
{{
  "reason": "<One sentence explanation on the rejection>",
  "decision": "reject"
}}
\end{Verbatim}
\end{tcolorbox}

\subsection{Diverse Prompt Sampling }
\begin{tcolorbox}[promptbox,title={Prompt for refining a problem description}]
\footnotesize
\begin{Verbatim}[breaklines=true,
  breakanywhere=true,
  breaksymbolleft=,
  breaksymbolright=]
You are an expert in constraint programming. Analyze the provided problem description and its corresponding input parameter specifications for logic and completeness. If you find any reasoning gaps, logical flaws, or missing information, revise and supplement the problem as needed based on your expertise.

After your analysis, remove any noise or irrelevant information, and rewrite the problem description to be concise and clear, making sure all essential details are explicitly stated. When referring to input parameters in your rewritten description, use the exact names given in the specifications-do not alter them.

Only provide a short analysis and the rewritten problem description. Do not provide the given input parameter specifications. Return your answer in the following JSON format:
{{
  "analysis": "xxx",
  "refined_description": "xxx"
}}

For example, given this original NQueens problem description and parameter specifications:
Problem description:
"Can $n$ queens (of the same colour) be placed on a $n\\times n$ chessboard so that none of the  queens can attack each other?\n\nIn chess a queen attacks other squares on the same row, column, or either diagonal as itself. So the $n$-queens problem is to find a set of $n$ locations on a chessboard, no two of which are on the same row, column or diagonal."
Input parameter specifications:
n:"The size of the chessboard and the number of queens, given as an integer."

Rewritten:
Place n queens on an n by n chessboard so that:
- No two queens are on the same row.
- No two queens are on the same column.
- No two queens are on the same diagonal (both upward and downward).

Now analyze the following problem context and generate an enhanced problem description:
Problem description:
{Problem description}
Input parameters specification:
{Specification of the input parameters}
\end{Verbatim}
\end{tcolorbox}

\begin{tcolorbox}[promptbox,title={Prompt for synthesizing a modeling plan}]
\footnotesize
\begin{Verbatim}[breaklines=true,
  breakanywhere=true,
  breaksymbolleft=,
  breaksymbolright=]
You are an expert in constraint programming. Your task is to break down a given CP problem into a sequence of sub-tasks, first with natural language descriptions, and then (in a later step) implementing them one by one in MiniZinc. You will be provided with a problem description, specification of the input parameter and expected output formats. The solution output will be formed based on the solution of the model in a later stage; for this task, do not design any task for displaying the solution or decision variable values in the model.
For each problem description:
1. Identify the main constraints required to model the problem in MiniZinc.
2. Decompose these constraints into sub-tasks. Each sub-task may address one or more logically related constraints.
3. List each sub-task concisely in natural language, which will be addressed sequentially. Note:
   - The first sub-task should always be: "load the given parameters"
   - Wrap each sub-task description in tags of the form `<task{{id}}> ... </task{{id}}>`, where `{{id}}` is the task number (e.g., `<task1> ... </task1>`).

**Example**: 
For example, for the N-Queens problem: 
- problem description:
"Can n queens be placed on a n by n chessboard so that no two Queens are on the same row, column or diagonal."
- details of the data_dict:
"n": "The size of the chessboard and the number of queens, given as an integer."
- output solutions and formats
(1) `queens`: An array representing the positions of the queens on the chessboard. The value at each index `i` represents the row position of the queen in the ( i )-th column.","size": "[n]"


**Breakdown**:  
<task1>Task 1: Load Parameters and Initialize Decision Variables 
- Load the board size (n) from the input parameter.
- Define an array of n decision variables, queens[1..n], where each variable represents the column position of a queen in a distinct row. This structure enforces that each queen automatically occupies a unique row.
</task1>

<task2>Task 2: Enforce Column Constraints  
- Ensure that no two queens share the same column by using an all_different constraint on the queens array.
</task2>

<task3>Task 4: Enforce Diagonal Constraints
- Ensure that no two queens share the same diagonal (both main and anti-diagonals). 
</task3>

<task4>Task 4: Solve the model
- Add `solve satisfy;` to invoke the solver.
</task4>

Now, analyze the following problem context, generate the step-by-step modeling strategy:
Problem description:
{Problem description}

Input parameters:
{Specification of the input parameters}

Required output formats:
(Reminder: Do not design any task for displaying or printing outputs.)
{Specification of the required output formats}
\end{Verbatim}
\end{tcolorbox}

\end{document}